\pdfoutput=1
\documentclass[runningheads]{llncs}
\usepackage{graphicx}
\usepackage{multirow}
\usepackage{bm}
\usepackage{amsmath}
\usepackage{amssymb}
\usepackage{algorithm}
\usepackage{algorithmic}
\usepackage{booktabs}
\usepackage{bbding}
\usepackage[breaklinks,colorlinks,citecolor=blue]{hyperref} 
\usepackage{wrapfig}
\usepackage[labelfont=bf, labelsep=period]{caption}
\usepackage{xcolor}
\usepackage[normalem]{ulem}

\begin{document}
\title{WikiIns: A High-Quality Dataset for Controlled Text Editing by Natural Language Instruction}
\titlerunning{WikiIns}
%
\author{Xiang Chen\inst{1,2,3} \and
Zheng Li\inst{1,3,4} \and
Xiaojun Wan\inst{1,2,3}\thanks{Corresponding author} }
\authorrunning{Chen et al.}

\institute{Wangxuan Institute of Computer Technology, Peking University \and
Center for Data Science, Peking University 
\and 
The MOE Key Laboratory of Computational Linguistics, Peking University \and
Schools of Electronic Engineering and Computer Science, Peking University \\
\email{\{caspar,lizheng2001,wanxiaojun\}@pku.edu.cn}}
\maketitle              
\begin{abstract}
Text editing, i.e., the process of modifying or manipulating text, is a crucial step in human writing process. In this paper, we study the problem of \textit{controlled text editing} by \textit{natural language instruction}. According to a given instruction that conveys the edit intention and necessary information, an original draft text is required to be revised into a target text. Existing automatically constructed datasets for this task are limited because they do not have informative natural language instruction. The informativeness requires the information contained in the instruction to be enough to produce the revised text. To address this limitation, we build and release WikiIns, a high-quality controlled text editing dataset with improved informativeness. We first preprocess the Wikipedia edit history database to extract the raw data (WikiIns-Raw). Then we crowdsource high-quality validation and test sets, as well as a small-scale training set (WikiIns-Gold). With the high-quality annotated dataset, we further propose automatic approaches to generate a large-scale ``silver'' training set (WikiIns-Silver). Finally, we provide some insightful analysis on our WikiIns dataset, including the evaluation results and the edit intention analysis. Our analysis and the experiment results on WikiIns may assist the ongoing research on text editing. The dataset, source code and annotation guideline are available at \url{https://github.com/CasparSwift/WikiIns}.

\end{abstract}

\section{Introduction}

Text editing has recently become a new scenario for text generation. Previous works~\cite{malmi-etal-2019-encode,stahlberg2020seq2edits,mallinson2020felix,iso2020fact} have proposed several text editing architectures for downstream text generation tasks such as text simplification, sentence fusion, and grammatical error correction. 



In this study, we aim to construct a high-quality text editing benchmark to better evaluate the text editing models. Previous works typically evaluate their methods for a particular type of edit, such as text simplification or grammar correction. However, it is inadequate to evaluate methods only on specific edits because there are various categories for human editing~\cite{yang2017identifying}. Considering that natural language is the simplest way to express all kinds of edit operations, in this work, we focus on the task of \textit{controlled text editing} by \textit{natural language instruction}. The given instructions typically express the intention of edits and also convey the necessary information for editing to the users.


Wikipedia edit history~\cite{nunes2008wikichanges} is a natural resource to construct the benchmark for controlled text editing by natural language instruction, which contains the old (\textit{draft text}) and new (\textit{revised text}) versions of Wikipedia documents and the comments written by the editors. Existing datasets~\cite{zhang2019modeling,faltings2021text} follow the automatic data mining process, which directly views the editors' comments as a proxy for the natural language instruction. However, this automatic way will inherently introduce too much noise. For instance, as illustrated in Table~\ref{tab:comparsion1}, the instruction ``reword'' fails to give enough information for the word insertions (e.g., ``user profiles'', ``messaging'', ``commenting''). It is infeasible to reproduce the target only given the source and instruction. This case belongs to open-ended text editing tasks rather than controlled text editing tasks, and evaluation is very hard for open-ended text editing tasks. 

\begin{table}[t!]
    \centering
    \caption{A simple illustration of the difference between WikiDocEdits and our proposed WikiIns. The \textcolor{red}{words in red} are the drastically changed content. The \textcolor{orange}{words in orange} are the content to be deleted. The \textcolor{blue}{words in blue} are the content to be inserted. Our instructions are more informative.}
    \label{tab:comparsion1}
    \resizebox{0.78\textwidth}{!}{
    \begin{tabular}{p{1.0\columnwidth}}
    \toprule
    \textbf{WikiDocEdits}~\cite{faltings2021text} \\
    \midrule
    \textbf{[Instruction]} Reword \\
    \textbf{[Source]} ByteDance responded by adding a kids-only mode to TikTok which \textcolor{red}{allows music videos to be recorded, but not posted and by removing some accounts and content from those determined to be underage}.\\
    \textbf{[Target]} ByteDance responded by adding a kids-only mode to TikTok which \textcolor{red}{blocks the upload of videos, the building of user profiles, direct messaging, and commenting on other’s videos, while still allowing the viewing and recording of content}. \\
    \midrule[1pt]
    \textbf{WikiIns} (this work) \\
    \midrule
    \textbf{[Instruction]} Apple USB modem no longer a current model \\
    \textbf{[Source]} Apple also sells a variety of computer accessories for Mac computers including the AirPort wireless networking products, Time Capsule, Cinema Display, Magic Mouse, Magic Trackpad, Wireless Keyboard, the Apple Battery Charger \textcolor{orange}{and the Apple USB Modem}. \\
    \textbf{[Target]} Apple also sells a variety of computer accessories for Mac computers including the AirPort wireless networking products, Time Capsule, Cinema Display, Magic Mouse, Magic Trackpad, Wireless Keyboard, \textcolor{blue}{and} the Apple Battery Charger. \\
    \bottomrule
    \end{tabular}
    }
    \vspace{-5mm}
\end{table}

To address this problem, we propose WikiIns, a high-quality controlled text editing dataset with informative natural language instruction. Table~\ref{tab: comparison of datasets} demonstrates the differences between WikiIns and existing datasets. In this paper, we design a reproducible methodology to improve informativeness. First, based on a small subset of the corpus, we employ human annotators to manually filter the noisy samples and rewrite the instructions that are not informative. In this way, we crowdsource high-quality validation and test datasets, as well as a relatively small training dataset for controlled text editing. Then we use a model trained on this small dataset to automatically filter the noisy samples on the entire corpus, which produces a large-scale silver training set. Figure~\ref{fig: method} presents an overview of the data process methodology.

\begin{table}[t!]
    \centering
\begin{minipage}[h]{0.52\textwidth}
    \captionsetup{type=figure}
    \caption{An overview of dataset construction procedure for proposed WikiIns.}
    \label{fig: method}
    \centering
    \includegraphics[width=1.0\textwidth]{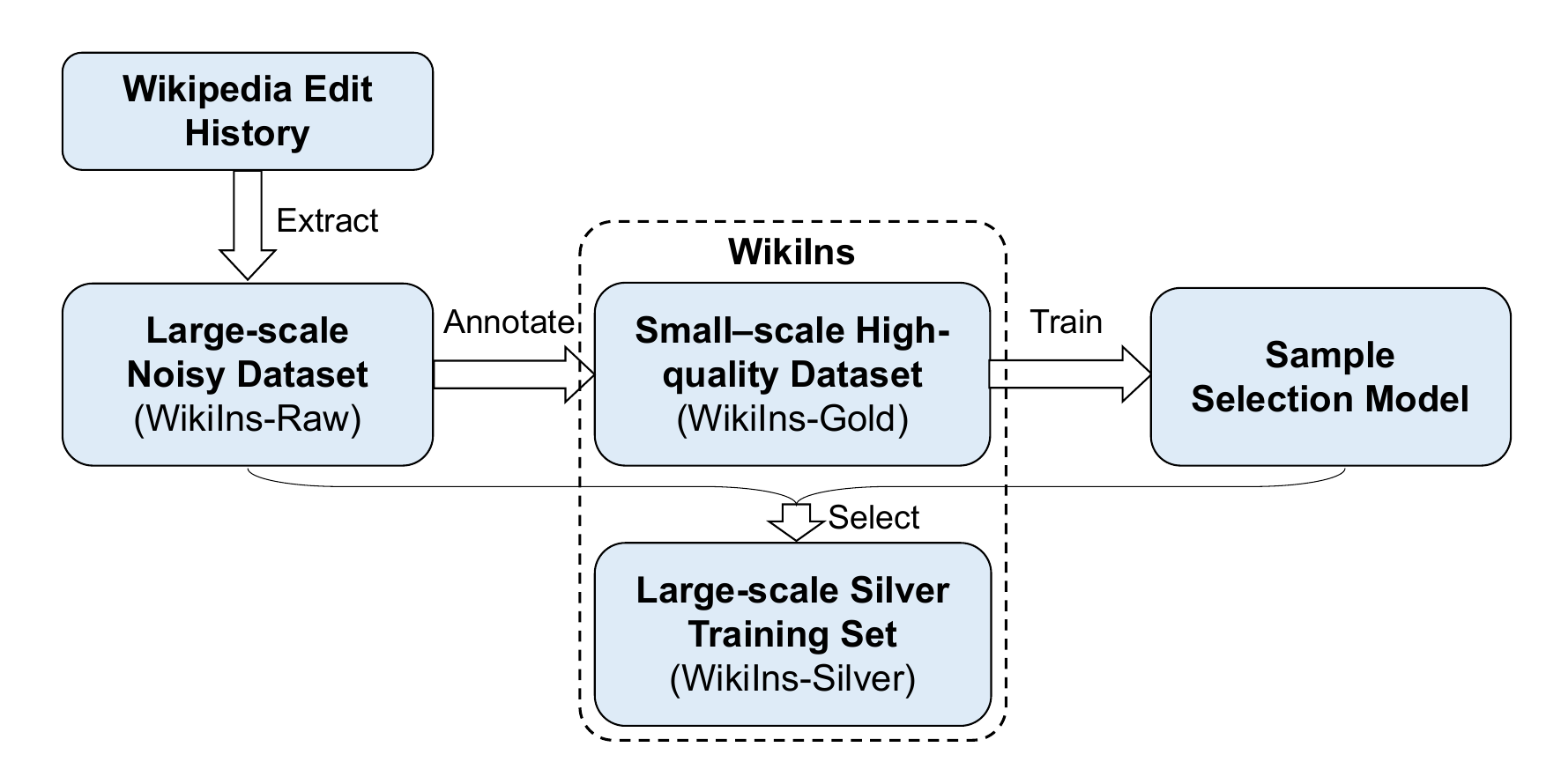}
\end{minipage}
\hfill
\begin{minipage}[h]{0.46\textwidth}
    \centering
    \caption{Comparisons with existing related datasets based on Wikipedia edit history. }
    \label{tab: comparison of datasets}
    \resizebox{0.98\textwidth}{!}{
    \begin{tabular}{l|c|c|c}
    \toprule
    Dataset & Instruction Type & Annotation & Informative  \\
    \midrule
    Yang et al. \cite{yang2017identifying} & Taxonomic & \Checkmark & \XSolidBrush  \\
    Faruqui et al. \cite{faruqui2018wikiatomicedits} & Atomic & \XSolidBrush & \Checkmark  \\
    Zhang et al. \cite{zhang2019modeling} & Natural Language & \XSolidBrush & \XSolidBrush  \\
    Iso et al. \cite{iso2020fact} & Fact Triples & \XSolidBrush & \Checkmark  \\
    Anthonio et al. \cite{anthonio2020wikihowtoimprove} & - & \XSolidBrush & \XSolidBrush  \\
    Faltings et al. \cite{faltings2021text} & Natural Language & \XSolidBrush & \XSolidBrush  \\
    Du et al. \cite{du2022understanding} & Taxonomic & \Checkmark & \XSolidBrush \\
    Spangher et al. \cite{spangher2022newsedits} & Taxonomic & \Checkmark & \XSolidBrush \\
    Jiang et al. \cite{jiang2022arxivedits} & Taxonomic & \Checkmark & \XSolidBrush \\
    WikiIns (this work) & Natural Language & \Checkmark & \Checkmark \\
    \bottomrule
    \end{tabular}
    }
\end{minipage}
\end{table}

Furthermore, we conduct extensive experiments and analyses on WikiIns, proving the quality improvement over previous datasets. We also implement some prevailing text editing models as well as some conventional seq2seq models as strong baselines. We believe that WikiIns is valuable for researchers to explore text editing models. Our contributions can be summarized as follows:
\begin{itemize}
    \item Propose and formulate the ``Controlled Text Editing'' as a standalone task.
    \item Build and release WikiIns, a dataset for controlled text editing with informative natural language instructions.
    \item Conduct extensive experiments and empirical analysis on WikiIns to provide an in-depth understanding of the dataset, which may aid future research for better text editing model design.
\end{itemize}

\section{Related Work}

\paragraph{Datasets for Text Editing}
Wikipedia edit history is an important resource to build text editing datasets. Yang et al. \cite{yang2017identifying} constructed a dataset to classify the edits of English Wikipedia into 13 categories of edit intentions. Faruqui et al. \cite{faruqui2018wikiatomicedits} focused on the atomic edits where the instruction is a single span of text. 
Zhang et al. \cite{zhang2019modeling} focused on predicting the locations of edits through the comments and identify the most relevant comment for an edit. Anthonio et al. \cite{anthonio2020wikihowtoimprove} extracted data from wikiHow and further investigated the edits by manual annotation. Faltings et al~\cite{faltings2021text} proposed a new text editing task that aims to edit text by the user's command and the grounding. However, as shown in Table~\ref{tab: comparison of datasets}, none of these works has human-annotated natural language instruction and follows the informativeness rule. To address these limitations, we propose WikiIns for controlled text editing with informative instructions.

\paragraph{Models for Text Editing}
Text editing models have recently attracted widespread attention. Gu et al.~\cite{gu2019levenshtein} proposed Levenshtein Transformer (LevT) with non-autoregressive decoding by insertion and deletion. Malmi et al.~\cite{malmi-etal-2019-encode} proposed LaserTagger to predict text editing operations over a fixed set. Mallinson et al.~\cite{mallinson2020felix} proposed FELIX, which contains two components for tagging and insertion. The tagging model is based on a BERT~\cite{Devlin2019BERTPO} encoder and pointer network~\cite{Vinyals2015PointerN}. The insertion model uses another BERT to predict masked tokens. Other general text editing models include Seq2Edits~\cite{stahlberg2020seq2edits}, CopySpan~\cite{Panthaplackel2021CopyTE}, and EDIT5~\cite{Mallinson2022EdiT5ST}.


\section{Problem Formulation}
For the task of controlled text editing investigated in this study, each data sample consists of: (1) Source: the source draft text $X$, (2) Instruction: an \textbf{informative} instruction for text editing $Z$, (3) Target: the revised text after editing operations $Y$. We formulate \textbf{Controlled Text Editing} as a task to generate the target $Y$, given $X$ and $Z$: $p(Y|X, Z;\theta) = \prod_{t=1}^{N_Y} p(y_t|y_{<t},X, Z; \theta)$,
where $\theta$ is the model parameters and $N_Y$ is the length of the target.


\section{Gold Dataset Creation}
\label{sec: gold dataset creation}

\subsection{Data Preprocessing}
\label{ssec: data preprocessing}

We follow the preprocessing process of~\cite{faltings2021text} to extract the (\textit{source, instruction, target}) triplets from the English Wikipedia dumps\footnote{The backup dumps can be downloaded at \url{https://dumps.wikimedia.org}}. We also use the WikiExtractor script to remove the HTML tags and Wikipedia markups. In addition, we find some heuristic rules to improve the informativeness of the comments. To make sure the comments are only for one single sentence, we omit documents with multiple edits. We also remove the samples with linguistically meaningless comments, such as modifying hyperlinks, citations, pictures, formats, and so forth. Aside from that, we notice that some copy-editings, such as edits about grammar, spelling, capitalization, and punctuation, can be easily identified by keyword matching for the comments. These edits do not require human annotation. After preprocessing, we obtain 1.17M samples with (\textit{source, instruction, target}) triplets. We name this corpus \textbf{WikiIns-Raw} in the following sections.

\subsection{Data Annotation}

\paragraph{Guideline}
We randomly select 20K samples from WikiIns-Raw and employ 20 annotators proficient in English to process the sampled data. Following the annotation guideline, the annotators are asked to make a judgment about \textbf{whether each instruction is informative}. Then, for the uninformative ones, we ask them to \textbf{rewrite the original instructions} to give informative instructions. Specifically, the annotation procedure follows the two-step approach as follows:

\begin{itemize}
    \item \textbf{Step 1:} Determine whether the instruction corresponds to the actual text edit. Samples with irrelevant instructions should be discarded or rewritten.
    \item \textbf{Step 2:} On the premise that the instruction can correctly describe the edit, further judge whether the instruction is informative enough to reproduce the revised text. The samples that we accept include (but are not limited to) directly pointing out how to edit the draft in the instruction, grammatical/spelling correction, and adding some factual information.
\end{itemize}

For instruction rewriting, we ask the annotators to avoid using trivial instructions, such as ``delete a word'' or ``add a word''. We also prefer slightly editing the original instructions rather than writing brand-new instructions. Furthermore, to help the annotators thoroughly understand the annotation rules,
we provide various examples in our annotation guideline. This may also alleviate the potential for annotation bias~\cite{parmar2022don} in the crowdsourcing process. 

\paragraph{Annotator Training}
We provided a training session before the annotation to select qualified annotators. During the training session, each annotator is shown the annotation guideline and asked to annotate 100 trial instances. We carefully examine the annotation quality and give some feedback to help them refine the annotation. Out of the initial 20 annotators, there are 13 annotators who complete the training session and continue to annotate the rest of the samples. We name this annotated corpus \textbf{WikiIns-Gold} in the following sections. WikiIns-Gold contains 6,060 samples with high-quality informative natural language instructions. There are 1,584 (26.1\%) instructions being rewritten in WikiIns-Gold.

\paragraph{Dataset Split}
We split the annotated data into training/validation/test subsets. To reduce the potential for content overlap among different subsets, we split the dataset at the document level. We ensure that any two samples from the same Wikipedia page\footnote{It's possible that two samples are within the same page because one single page may have multiple historical versions.} will be grouped into the same subset. Consequently, we obtain 4,060/1,000/1,000 samples as training, validation, and test sets, respectively.

\section{Silver Training Dataset Creation}

After data annotation, we get a relatively small dataset for training. However, it may be not enough to train supervised text editing models. Additionally, we fail to make full use of WikiIns-Raw in Section~\ref{sec: gold dataset creation} because we only select 20K samples from WikiIns-Raw (1.17M samples) for annotation. To obtain a much larger training set, we implement an automatic approach to select informative samples from WikiIns-Raw to construct a high-quality silver training dataset.


To select high-quality samples, we use a binary classifier to identify the informative samples. We first construct the training set for the sample classifier. All samples in the training set of WikiIns-Gold can be viewed as informative samples. The samples discarded or rewritten by the annotators can be viewed as non-informative samples. Therefore, we obtain a dataset including 18,554 samples with 4,060 informative samples. We concatenate the instruction, source, and target as the input. We choose T5$_{\text{base}}$~\cite{raffel2020exploring} as the backbone because it can handle such long inputs. After that, we train five binary classification models with five different random seeds and train-test splits. All these models vote to obtain the final prediction results. The average F1 score of these classifiers is 92.68 ($\pm$1.58).

After selection, we obtain a subset of 337,363 samples from WikiIns-Raw. We name this corpus \textbf{WikiIns-Silver} in the following sections. Our proposed \textbf{WikiIns} is the combination of \textbf{WikiIns-Gold} and \textbf{WikiIns-Silver}.

\begin{table}[t!]
\centering
\begin{minipage}[t]{0.48\textwidth}
\centering
\caption{WikiIns-Gold statistics.}
\label{tab: statistics}
\resizebox{0.98\textwidth}{!}{
\begin{tabular}{lccccc}
\toprule
\multirow{2}{*}{Statistics} & \multicolumn{3}{c}{Percentiles} & \multirow{2}{*}{Max} & \multirow{2}{*}{Mean}  \\ \cmidrule(lr){2-4} 
    & 25\% & 50\% & 75\% &     & \\
\midrule
Instruction Length   & 3    & 6    & 12   & 89  & 8.85  \\
Source Length         & 20   & 29   & 39   & 178 & 31.34 \\
Target Length  & 21   & 29   & 39   & 177 & 31.44 \\
Levenshtein Distance & 1    & 1    & 3    & 38  & 2.60  \\
\bottomrule
\end{tabular}
}
\end{minipage}%
\begin{minipage}[t]{0.48\textwidth}
\centering
\caption{WikiIns-Silver statistics.}
\label{tab: statistics-silver}
\resizebox{0.98\textwidth}{!}{
\begin{tabular}{lccccc}
\toprule
\multirow{2}{*}{Statistics} & \multicolumn{3}{c}{Percentiles} & \multirow{2}{*}{Max} & \multirow{2}{*}{Mean} \\ \cmidrule(lr){2-4}
                     & 25\% & 50\% & 75\% &     &       \\ 
\midrule
Instruction Length   & 3 & 6 & 13 & 112 & 9.31 \\
Source Length         & 21 & 30 & 41 & 1362 & 33.20 \\
Target Length  & 21 & 30 & 41 & 1362 & 33.14  \\
Levenshtein Distance & 1 & 1 & 2 & 114 & 2.31  \\ 
\bottomrule
\end{tabular}
}
\end{minipage}
\vspace{-4mm}
\end{table}

\section{Dataset Analysis}

\subsection{Statistics}
We provide an overview of WikiIns in Tables~\ref{tab: statistics} and~\ref{tab: statistics-silver}. We calculate the 25\%, 50\%, 75\% percentiles, maximum, and mean values of instruction/source/target length, and the levenshtein distance between the source and target. The length denotes the number of words. The levenshtein distance~\cite{levenshtein1966binary} is calculated at the word level. The statistics show that the instruction is typically a short sentence or phrase, and shorter than the source and target text. Moreover, most samples only require a modification of no more than three words. Note that we omit some extremely long sentences to prevent long inputs.

\begin{table}[t!]
\centering
\small
\caption{The distribution of edit intention. ``Y17'' indicates the distribution of the dataset proposed by Yang et al.~\cite{yang2017identifying}. ``F21'' indicates the distribution for WikiDocEdits~\cite{faltings2021text}, which is also predicted by the classifier of Yang et al.~\cite{yang2017identifying}. ``F21'' also shares similar data distribution with WikiIns-Raw due to the similar data construction process. The percentages do not total 100 because some edits may have multiple labels.}
\label{tab:group}
\resizebox{0.9\textwidth}{!}{
\begin{tabular}{l|p{0.5\columnwidth}|cccc}
\toprule
\multirow{2}{*}{Group} & \multirow{2}{*}{Edit Intention Label} & 
\multicolumn{1}{c}{\multirow{2}{*}{Y17}} &
\multicolumn{1}{c}{\multirow{2}{*}{F21}} &
\multicolumn{1}{c}{\multirow{2}{*}{WikiIns-Gold}} & 
\multicolumn{1}{c}{\multirow{2}{*}{WikiIns-Silver}} \\ 
&  &  &  &  \\
\midrule
Fluency & Refactoring, Copy Editing, Wikification, Point of View & 61.45 & 57.00 & 87.06 & 90.23 \\
Content & Fact Update, Simplification, Elaboration, Verification, Clarification & 38.74 & 24.77 & 17.41 & 11.35\\
Other & Vandalism, Counter Vandalism, Process, Disambiguation & 14.37 & 26.65 & 2.64 & 2.73 \\
\bottomrule
\end{tabular}
}
\vspace{-5mm}
\end{table}

\subsection{Edit Intention Distribution}

In this section, we aim to analyze the distribution of the edit intentions on WikiIns-Gold and WikiIns-Silver, as many previous works have done. Yang et al.~\cite{yang2017identifying} proposed a taxonomy of edit intentions that contains 13 categories and released a dataset with human-annotated edit intentions. 
We use their script to extract features and apply their classifier to WikiIns.

Similar to~\cite{faltings2021text}, we divide the 13 categories of edits into three groups: ``Fluency'', ``Content'', and ``Other''. The group ``Fluency'' includes some simple edits for improving text fluency. The group ``Content'' includes the edits that add/delete specific content and information. The group ``Other'' includes vandalism, counter-vandalism, and some Wikipedia-specific edits, which lack linguistic information. The group ``Other'' is not so important because we focus on the syntactic or semantic edits in this paper. 

Table~\ref{tab:group} reports the predicted edit intention distributions of WikiIns-Gold and WikiIns-Silver. It can be observed that most of the edits can be categorized into the group ``Fluency''. The proportion of the group ``Other'' decreases compared to~\cite{yang2017identifying} and~\cite{faltings2021text} because human annotation has removed most of these edits. The proportion of the group ``Content'' also decreases because many edits of this group are not controllable. The human annotation tends to remove these edits to better fit the task setting of controlled text editing. 



\begin{table*}[t!]
    \centering
    \small
    \caption{Examples with different types of instructions from WikiIns. The \textcolor{orange}{words in orange} are the content to be deleted. The \textcolor{blue}{words in blue} are the content to be inserted. The four types are ranked according to their difficulties. It should be noted that WikiIns contains various instructions, which are not limited to these four types.}
    \label{tab:examples}
    \resizebox{0.9\textwidth}{!}{
    \begin{tabular}{l|p{0.25\columnwidth}|p{0.7\columnwidth}}
    \toprule
    Type & Instruction & Edits \\
    \midrule 
    \multirow{2}{*}{\uppercase\expandafter{\romannumeral1}} & Changed ``'60s'' text to ``1960s'' & Aside from a few minor hits in Australia, he also failed to produce any charting singles after the \textcolor{orange}{\sout{'60s}} \textcolor{blue}{1960s}. \\ 
    & Reword ``his green card'' with ``permanent residency'' & The following year, his US immigration status finally resolved, Lennon received \textcolor{orange}{\sout{his green card}} \textcolor{blue}{permanent residency}, and when Jimmy Carter was inaugurated as president in January 1977, Lennon and Ono attended the Inaugural Ball. \\
    \midrule
    \multirow{2}{*}{\uppercase\expandafter{\romannumeral2}} & Copyedit: spelling. & British nationalists have long \textcolor{orange}{\sout{campagined}} \textcolor{blue}{campaigned} against EU \textcolor{orange}{\sout{intergration}} \textcolor{blue}{integration}. \\
    & Grammar issue & Reparations would also go towards the reconstruction costs in other countries, including Belgium, which \textcolor{orange}{\sout{was}} \textcolor{blue}{were} also directly affected by the war. \\
    \midrule
    \multirow{1}{*}{\uppercase\expandafter{\romannumeral3}} & Delete the content in the brackets & Early research suggested that injection of lethal quantities of astatine caused morphological changes in breast tissue \textcolor{orange}{\sout{(although not other tissues)}}; this conclusion remains controversial. \\
    \midrule 
    \multirow{2}{*}{\uppercase\expandafter{\romannumeral4}} & ``While'' implies ``simultaneously'' &  Returning from the US in January 1973, they recorded ``Brain Damage'', ``Eclipse'', ``Any Colour You Like'' and ``On the Run'', while \textcolor{orange}{\sout{simultaneously}} fine-tuning the work they had already laid down in the previous sessions. \\
    & Add the info: there are 700 employees in the facility & The William Wrigley Company has a large manufacturing facility in Gainesville \textcolor{blue}{with 700 employees}. \\
    \bottomrule
    \end{tabular}
    }
    \vspace{-10mm}
\end{table*}

\subsection{Examples of WikiIns}
\label{ssec: examples of wikiins}
Table~\ref{tab:examples} shows some examples of WikiIns. The examples are presented to illustrate the different types of instructions as well as different levels of difficulty for our dataset. The four types are ordered by their difficulties. Type \uppercase\expandafter{\romannumeral1} is usually a straightforward phrase substitution. Type \uppercase\expandafter{\romannumeral2} is grammar or spelling correction. Type \uppercase\expandafter{\romannumeral1} and Type \uppercase\expandafter{\romannumeral2} are relatively easier to learn for text editing models. Type \uppercase\expandafter{\romannumeral3} is more complex, which requires a better understanding of the meaning of the instruction to infer which part should be edited. Type \uppercase\expandafter{\romannumeral4} is the most difficult, which requires the ability to extract the potential edit intention of the instruction as well as some reasoning ability.

\section{Experiment}

In this section, we conduct extensive experiments to investigate the following research questions:

RQ1: Do WikiIns-Gold and WikiIns-Silver improve informativeness compared to WikiIns-Raw?

RQ2: How well do the different text editing models perform on WikiIns?


\subsection{RQ1: Informativeness Improvement}

\paragraph{Evaluation Metrics}
We adopt multiple automatic metrics to evaluate the generation quality of controlled text editing:

\begin{itemize}
    \item \textbf{Exact Match (EM)}: the proportion of correct edit. An edit is correct if and only if the whole sentence is exactly the same as the reference.
    \item \textbf{SARI}~\cite{xu-etal-2016-optimizing}: the average of \textbf{KEEP}, \textbf{ADD}, and \textbf{DEL} score, which are the F1 score for keep/add/delete operations, respectively.
    \item \textbf{BLEU}~\cite{papineni2002bleu}: a natural choice but not very reliable because of highly overlapped inputs and outputs. We provide the BLEU anyway, for reference only.
    \item \textbf{Word Edit}~\cite{faltings2021text}: the precision, recall, and F1 score based on word edit sets\footnote{See the Appendix of~\cite{faltings2021text} for details.}. 
\end{itemize}

\begin{table*}[t!]
\centering
\caption{Evaluation results for controlled text editing by instruction on WikiIns test set. ``\#N'': the number of training data. ``w.o. ins'': training without instruction. ``random'': training with a random subset of WikiIns-Raw. ``w.o. rewritten'': using the original instructions, not the instructions rewritten by the annotators.}
\label{tab:main result}
\resizebox{0.97\textwidth}{!}{
\begin{tabular}{@{}lccrrrrrrrrr@{}}
\toprule
\multirow{2}{*}{Method} &
  \multicolumn{1}{c}{\multirow{2}{*}{Training Set}} &
  \multicolumn{1}{c}{\multirow{2}{*}{\#N}} &
  \multicolumn{1}{c}{\multirow{2}{*}{EM}} &
  \multicolumn{1}{c}{\multirow{2}{*}{SARI}} &
  \multicolumn{1}{c}{\multirow{2}{*}{KEEP}} &
  \multicolumn{1}{c}{\multirow{2}{*}{ADD}} &
  \multicolumn{1}{c}{\multirow{2}{*}{DEL}} &
  \multicolumn{1}{c}{\multirow{2}{*}{BLEU}} &
  \multicolumn{3}{c}{Word Edit} \\ \cmidrule(l){10-12} 
 &
  \multicolumn{1}{c}{} &
  \multicolumn{1}{c}{} &
  \multicolumn{1}{c}{} &
  \multicolumn{1}{c}{} &
  \multicolumn{1}{c}{} &
  \multicolumn{1}{c}{} &
  \multicolumn{1}{c}{} &
  \multicolumn{1}{c}{} &
  \multicolumn{1}{c}{P} &
  \multicolumn{1}{c}{R} &
  \multicolumn{1}{c}{F1} \\ \midrule
Copying Source & WikiIns-Gold & - & 0.00 & 50.29 & 97.82 & 28.23 & 24.82 & 89.85 & 0.00 & 0.00 & 0.00  \\ \midrule
T5$_{\text{base}}$ (w.o. ins) & WikiIns-Gold & 4060 & 9.20 & 53.71 & 97.60 & 32.54 & 31.00 & 89.10 & 24.91 & 9.04 & 13.26 \\
T5$_{\text{base}}$ (random) & WikiIns-Raw & 4060 & 17.40 & 59.71 & 97.64 & 41.85 & 39.63 & 89.41 & 28.19 & 15.88 & 20.31 \\
T5$_{\text{base}}$ (w.o. rewritten) & WikiIns-Gold & 4060 & 27.40 & 65.84 & 97.72 & 50.24 & 49.56 & 90.45 & \bf 40.31 & 24.61 & 30.56 \\
T5$_{\text{base}}$ & WikiIns-Gold & 4060 &\bf 30.80 & \bf 68.38 & \bf 98.15 & \bf 54.03 & \bf 52.96 & \bf 90.91 &  38.65 & \bf 27.11 & \bf 31.87 \\ \bottomrule
\end{tabular}
}
\vspace{-2mm}
\end{table*}

\paragraph{Results on WikiIns-Gold} We train multiple models with WikiIns-Gold and evaluate these models on the test set and compare them with several baselines (simply copying the source, training without instruction, and using WikiIns-Raw). Table~\ref{tab:main result} provides the experiment results. Our findings are three-fold: (1) Training with WikiIns-Gold produces substantial improvement over training with equal size of WikiIns-Raw, which illustrates that our data annotation improves the data quality for controlled text editing tasks. (2) Removing the instruction leads to performance degradation on all the metrics, which demonstrates the importance of the instruction for WikiIns, and proves that it is difficult for T5 model to learn shortcuts only from the source text. (3) BLEU is not so reliable for this task on WikiIns. Due to the highly overlapped inputs and outputs, outputting the source text without any editing can achieve high BLEU scores. The word-level metrics (e.g., SARI, Word Edit) may better reflect the actual performance.

\begin{figure*}[t!]
    \centering
    \includegraphics[width=0.98\textwidth]{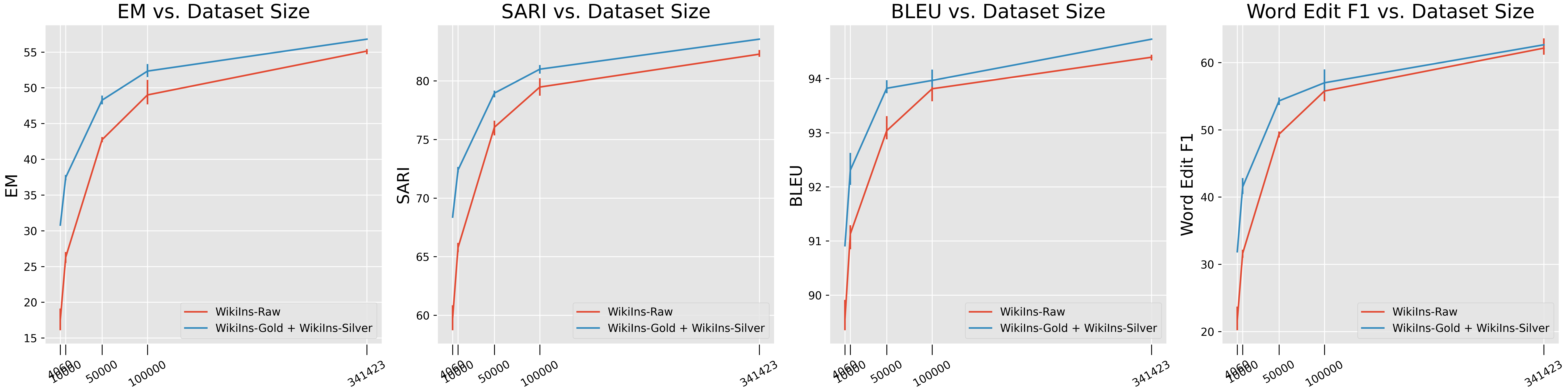}
    \caption{Evaluation results for different dataset sizes of training data. The error bars are computed by three runs with three different random seeds.}
    \label{fig: metric}
    \vspace{-2mm}
\end{figure*}

\begin{table}[t!]
    \centering
    \small
    \caption{Results of the BERTScore measurement $R_{\text{ins}}$, $R_{\text{del}}$, and $R_{\text{info}}$.}
    \label{tab:informativeness}
    \begin{tabular}{l|ccc}
    \toprule
     Dataset & $R_{\text{ins}}$ & $R_{\text{del}}$ & $R_{\text{info}}$ \\
     \midrule
     WikiIns-Raw & 0.3277 & 0.2857 & 0.3067 \\
     \midrule
     WikiIns-Silver & 0.4494 & 0.4248 & 0.4371 \\
     \midrule
     WikiIns-Gold (train) & \bf 0.4802 & 0.4268 & 0.4535 \\
     WikiIns-Gold (valid) & 0.4478 & 0.4311 & 0.4394 \\
     WikiIns-Gold (test) & 0.4798 & \bf 0.4621 & \bf 0.4710 \\
    \bottomrule
    \end{tabular}
    \vspace{-3mm}
\end{table}

\paragraph{Results on WikiIns-Silver} Furthermore, we train T5$_{\text{base}}$ with different sizes of WikiIns-Silver. We also use the training set of WikiIns-Gold. Figure~\ref{fig: metric} shows the results. 
We find that combing WikiIns-Gold and WikiIns-Silver has better \textit{sample efficiency} than WikiIns-Raw. With a relatively small dataset size, training with our dataset outperforms randomly selecting samples from WikiIns-Raw by a large margin. Moreover, training with WikiIns-Silver can achieve competitive performance compared to training on WikiIns-Raw with less training data, which reduces training costs. This result demonstrates the good quality and usefulness of WikiIns-Silver for controlled text editing tasks. 

\paragraph{BERTScore Measurement}
\label{ssec: quantitative} We further evaluate the informativeness improvement of WikiIns-Gold and silver data. We define the BERTScore~\cite{zhang2019bertscore} measurement of informativeness $R_{\text{info}}(X, Z, Y)=\frac12(R_{\text{ins}}+R_{\text{del}})$, where
\begin{small}
\begin{equation*}
R_{\text{ins}}=\frac{1}{|Y-X|}\sum_{x\in Y-X} \max_{z\in Z} f(x)^\top f(z),\quad R_{\text{del}}=\frac{1}{|X-Y|}\sum_{x\in X-Y} \max_{z\in Z} f(x)^\top f(z).
\end{equation*}
\end{small}
In this equation, $f(x)$ and $f(z)$ are the contextualized representations of the tokens $x$ and $z$ by BERT~\cite{Devlin2019BERTPO}, $Y-X=\{x|x\in Y\wedge x\notin X\}$ (inserted words), and $X-Y=\{x|x\in X\wedge x\notin Y\}$ (deleted words). We define $R_{\text{ins}}=0$ if $Y-X=\varnothing$, and $R_{\text{del}}=0$ if $X-Y=\varnothing$. We set $R_{\text{ins}}=R_{\text{del}}=R_{\text{info}}=1$ for instructions about grammar or spelling (i.e., Type \uppercase\expandafter{\romannumeral2} instruction described in Table~\ref{tab:examples}) because an instruction simply mentioning ``grammar'' or ``spelling'' is informative enough to reproduce the edits. We adopt keyword matching as an approximation to identify the instructions about grammar and spelling. We compute $R_{\text{info}}$ on both WikiIns and WikiIns-Raw. As shown in Table~\ref{tab:informativeness}, after the human annotation, there is a substantial improvement in informativeness compared to the raw data.

\begin{table*}[t!]
\centering
\small
\caption{Evaluation results of different text editing models on WikiIns test set. The training data includes both WikiIns-Gold training set and WikiIns-Silver.}
\label{tab: text editing result}
\begin{tabular}{lrrrrrrrrr}
\toprule
\multirow{2}{*}{Method} &
  \multicolumn{1}{c}{\multirow{2}{*}{EM}} &
  \multicolumn{1}{c}{\multirow{2}{*}{SARI}} &
  \multicolumn{1}{c}{\multirow{2}{*}{KEEP}} &
  \multicolumn{1}{c}{\multirow{2}{*}{ADD}} &
  \multicolumn{1}{c}{\multirow{2}{*}{DEL}} &
  \multicolumn{1}{c}{\multirow{2}{*}{BLEU}} &
  \multicolumn{3}{c}{Word Edit} \\ \cmidrule(l){8-10} 
 &
  \multicolumn{1}{c}{} &
  \multicolumn{1}{c}{} &
  \multicolumn{1}{c}{} &
  \multicolumn{1}{c}{} &
  \multicolumn{1}{c}{} &
  \multicolumn{1}{c}{} &
  \multicolumn{1}{c}{P} &
  \multicolumn{1}{c}{R} &
  \multicolumn{1}{c}{F1} \\ \midrule
\textbf{Text Editing Models} & \multicolumn{9}{r}{}                                                  \\
LevT~\cite{gu2019levenshtein} & 17.60 & 62.04 & 97.43 & 40.15 & 48.54 & 80.62 & 8.27 & 18.63 & 11.45 \\
FELIX~\cite{mallinson2020felix} & 18.40 & 67.13 & 96.99 & 49.04 & 55.37 & 82.17 & 7.68 & 28.71 & 12.12 \\ \midrule
\multicolumn{10}{l}{\textbf{Seq2seq Models}} \\
Transformer~\cite{vaswani2017attention} & 29.90 & 69.60 & 98.21 & 51.76 & 58.84 & 83.00 & 16.37 & 34.17 & 22.14 \\
BART$_{\text{large}}$~\cite{lewis2020bart}  & 55.80 & \textbf{83.96} & \textbf{99.01} & 75.39 & \textbf{77.48} & 93.15 & 56.89 & 57.23 & 57.06 \\
T5$_{\text{base}}$~\cite{raffel2020exploring} & \textbf{58.30} & 83.94 & 98.91 & \textbf{76.59} & 76.32 & \textbf{94.78} & \textbf{68.65} & \textbf{58.13} & \textbf{62.95} \\ \bottomrule
\end{tabular}
\vspace{-5mm}
\end{table*}

\subsection{RQ2: Evaluation of Text Editing Models}
In this section, we provide some baseline results for controlled text editing based on the full training set of WikiIns. We choose two representative text editing models: \textbf{LevT}~\cite{gu2019levenshtein} and \textbf{FELIX}~\cite{mallinson2020felix}. Moreover, we adopt some conventional seq2seq models, including vanilla Transformer~\cite{vaswani2017attention} ($d=512, l=6$), as well as pretrained BART$_{\text{large}}$~\cite{lewis2020bart} and T5$_{\text{base}}$~\cite{raffel2020exploring} models.

Table~\ref{tab: text editing result} demonstrates the experiment results. We find that text editing models still have much room for improvement compared to seq2seq models. One possible reason is that WikiIns only requires a slight modification over the draft, but \textbf{LevT} and \textbf{FELIX} may insert too many irrelevant words in the wrong position in the sentence. Pretrained models, T5$_{\text{base}}$ and BART$_{\text{large}}$, can improve the result by a large margin. However, the controlled text editing task is still challenging when considering fine-grained metrics (EM and Word Edit).

\section{Conclusion}
In this paper, we propose a dataset named WikiIns for controlled text editing by natural language instruction. We provide an in-depth analysis for WikiIns to better understand the characteristics of the data. Moreover, we further conduct extensive experiments to verify the data quality and provide baseline results of some text editing models. We believe WikiIns and our findings in this paper will help future research into the better design of text editing models.

\section*{Acknowledgment}
This work was supported by National Key R\&D Program of China (2021YFF0901502), National Science Foundation of China (No. 62161160339), State Key Laboratory of Media Convergence Production Technology and Systems and Key Laboratory of Science, Technology and Standard in Press Industry (Key Laboratory of Intelligent Press Media Technology). We appreciate the anonymous reviewers for their helpful comments. Xiaojun Wan is the corresponding author.

%
%
%
\bibliographystyle{splncs04}
\bibliography{ref}

\end{document}